# Exploiting Multilingualism in Low-resource Neural Machine Translation via Adversarial Learning

Amit Kumar, Ajay Pratap, *Member, IEEE*, and Anil Kumar Singh

***Abstract*—Generative Adversarial Networks *(GAN)* offer a promising approach for Neural Machine Translation *(NMT)*. However, feeding multiple morphologically languages into a single model during training reduces the NMT's performance. In GAN, similar to bilingual models, multilingual NMT only considers one reference translation for each sentence during model training. This single reference translation limits the GAN model from learning sufficient information about the source sentence representation. Thus, in this article, we propose Denoising Adversarial Auto-encoder-based Sentence Interpolation *(DAASI)* approach to perform sentence interpolation by learning the intermediate latent representation of the source and target sentences of multilingual language pairs. Apart from latent representation, we also use the Wasserstein-GAN approach for the multilingual NMT model by incorporating the model generated sentences of multiple languages for reward computation. This computed reward optimizes the performance of the GAN-based multilingual model in an effective manner. We demonstrate the experiments on low-resource language pairs and find that our approach outperforms the existing state-of-the-art approaches for multilingual NMT with a performance gain of up to 4 BLEU points. Moreover, we use our trained model on zero-shot language pairs under an unsupervised scenario and show the robustness of the proposed approach.**

## I. Introduction

Neural machine translation *(NMT)* has established a number of cutting-edge benchmarks in machine translation tasks [1], [2]. It is an encoder-decoder framework based on a sequence-to-sequence prediction model where the encoder generates the context vector by taking a source-side word as input, and the decoder decodes this generated context vector into target sequences. Recent works ( [3]–[5]) have extended the NMT approach to support multilingual translation, i.e., training a single model that can translate between multiple languages. There are several reasons for shifting researchers' interest from bilingual to multilingual machine translation. First, single model training for large number of languages makes multilingual model more cost-effective than multiple bilingual models. Another advantage is transfer learning i.e., training of low-resource languages in combination with high resource languages improves the translation quality of low-resource languages [6]. One of the fine examples of transfer learning-based NMT under multilingual scenario is Zero-Shot Translation *(ZST)* [2]. ZST is an unsupervised approach in which pretrained translation models are tested on related unseen language pairs.

There has been a significant amount of works done on multilingual machine translation, with the majority of the works focusing on translation between language pairs with English as a prominent language [7]–[9]. However, very few works on non-English language pairs exist [2], [10]. One of the reasons for less multilingual systems on non-English language pairs is insufficient training data for such language pairs. Multilingual models' translation quality is not as good as bilingual models due to multiple languages, which increases the data sparseness. However, use of multiple languages during training give some good systems for zero resource languages [11].

Multilingual training significantly improves the translation quality of low and zero-resource languages. Although fitting multiple morphologically rich language pairs into a single model suffers from representation learning bottlenecks and affects the generalisation capabilities, limiting the benefits of multilingualism on translation quality [12]. Despite a large amount of data fed into models with many parameters, translation performance in all language directions has not improved due to representation learning bottlenecks problems. Therefore, more research is needed for better data selection and representation, network architectures, and learning algorithms in low-resource multilingual model.

To solve the problems of representation learning and generalisation, we present Denoising Adversarial Auto-encoder-based Sentence Interpolation (*DAASI*) approach for low-resource multilingual machine translation. DAASI exploits the monolingual data by adversarial-based denoising autoencoder and generates the augmented data for parallel corpus. A denoising autoencoder is a neural network architecture that corrupts data and attempts to recreate it from corrupted samples, allowing it to perform well even when the inputs are noisy [13]. Then it uses sentence interpolation to generate the latent representation of data between two different languages and uses the interpolated data to train the NMT model relying on Generative Adversarial Network*(GAN)* [14]. We train the NMT based on Wasserstein Generative Adversarial Network (WGAN) [15], consists of generator and critic,

A. Kumar, A. Pratap, and A. K. Singh are with the Department of Computer Science and Engineering, Indian Institute of Technology (Banaras Hindu University) Varanasi 221005 India. E-mail: {amitkumar.rs.cse17, ajay.cse, aksingh.cse}@iitbhu.ac.in.

on newly constructed interpolated data. In generator, pre-trained NMT model produces translated sentence given a source sentence, while critic model takes sentence pairs as input, tries to learn the Wasserstein distance between them and judge whether they are real or fabricated based on the distances between the sentences. Unlike the reward computed in the regular GAN model, DAASI generates the reward on each test set of multiple languages.

We conducted the experiments on five low-resource language pairs: Gujarati (GU)↔Hindi (HI), Nepali (NE)↔Hindi (HI), Punjabi (PA)↔Hindi (HI), Maithili (MAI)↔Hindi (HI), Urdu (UR)↔Hindi (HI) to demonstrate the robustness of the proposed DAASI approach. We also use our pre-trained multilingual model on zero-shot language pairs (Bhojpuri (BHO)↔Hindi (HI) and Magahi (MAG)↔Hindi (HI)) under an unsupervised scenario. Particularly, the contributions of the paper are summarised as follows:

1) Propose *DAASI* approach based on denoising adversarial auto-encoder for low-resource multilingual machine translation.
2) Perform sentence interpolation on source-target language pairs to generate the intermediate latent representation of sentence that covers the diverse context of sentences in different languages.
3) Optimise the GAN for the multilingual machine translation model by incorporating WGAN and multi-language references to compute the reward.
4) Proposed approach outperforms existing state-of-the-art techniques up to 4 BLEU points in all translation tasks.

The rest of the paper is organized as follows. Section II discusses the closely related works. Section III describes the proposed solutions. Corpus statistics, the experimental setup and results conducted are reported in Section IV. Finally, paper is summarized in Section V with future aspects of work.

## II. Related Works

Limited availability of resources for low-resource machine translation could be improved under a multilingual training scenario. Projecting multiple languages in the same dimensional space helps the NMT systems utilize information from high-resource language pairs and enhance the translation quality of low-resource languages through a transfer learning approach. Recent studies on denoising autoencoders reveals few improvements in the NMT model under multilingual scenario. In this section, we closely review the multilingual machine translation systems in terms of denoising autoencoder.

### *A. Multilingual NMT without denoising autoencoder*

The main objective of multilingual NMT is to build a model that can assist translation between more than one language pair. Multilingual NMT is of three types: one-to-many [3], many-to-one [4] and many-to-many [5]. The learning objective for multilingual NMT is to maximize the log-likelihood of all training examples for all language pairs.

Table I: Comparison of existing works

| **Paper** | **A** | **B** | **C** | **D** | **E** | **F** |
|---|---|---|---|---|---|---|
| [3] | ✗ | ✗ | ✗ | ✓ | ✗ | ✗ |
| [4] | ✗ | ✗ | ✗ | ✓ | ✗ | ✗ |
| [5] | ✗ | ✗ | ✗ | ✓ | ✗ | ✗ |
| [7] | ✗ | ✗ | ✗ | ✓ | ✗ | ✗ |
| [8] | ✗ | ✗ | ✗ | ✓ | ✗ | ✗ |
| [9] | ✗ | ✗ | ✗ | ✓ | ✗ | ✗ |
| [16] | ✗ | ✗ | ✗ | ✓ | ✗ | ✗ |
| [17] | ✗ | ✗ | ✗ | ✓ | ✗ | ✗ |
| [18] | ✗ | ✗ | ✗ | ✓ | ✗ | ✗ |
| [19] | ✗ | ✗ | ✗ | ✓ | ✗ | ✗ |
| [20] | ✗ | ✗ | ✗ | ✓ | ✗ | ✗ |
| [21] | ✗ | ✗ | ✗ | ✓ | ✗ | ✗ |
| [22] | ✗ | ✗ | ✗ | ✓ | ✗ | ✗ |
| [23] | ✓ | ✗ | ✗ | ✗ | ✗ | ✗ |
| [24] | ✓ | ✗ | ✗ | ✓ | ✗ | ✗ |
| [25] | ✓ | ✗ | ✗ | ✗ | ✗ | ✗ |
| [26] | ✗ | ✗ | ✗ | ✗ | ✗ | ✗ |
| [DAASI] | ✓ | ✓ | ✓ | ✓ | ✓ | ✓ |

Note– **A**: Denoising Auto-Encoder, **B**: Wasserstein Generative Adversarial Network, **C**: WX [27], **D**: Multilingual, **E**: Sentence interpolation, **F**: Multi-language reward.

In [7], authors proposed architecture for NMT by incorporating an intermediate attention bridge between all languages and receiving multilingual sentence representations. In [8], authors conducted experiments on training massively multilingual NMT models, incorporating up to 103 different languages and 204 translation directions at the same time, and investigated various setups for training such models as well as analysing the trade-offs between translation quality and different modelling decisions. In [16], authors employed multilingual and multi-way neural machine translation approaches for morphologically rich languages, such as Estonian and Russian. In [17], authors addressed the rare word problem in multilingual MT models for low-resource language pairs.

In [9], authors proposed a multilingual lexicon encoding framework designed exclusively to smartly share lexical-level information without involving any heuristic data preprocessing. In [18], authors incorporated a language-aware interlingua into the Encoder-Decoder architecture. The incorporated interlingual network enables the model to learn a language-independent representation from the semantic spaces of different languages. In [19], authors investigated methods for improving massively multilingual NMT, particularly on ZST, and demonstrated that multilingual NMT has limited capacity, which they propose to improve by deepening the Transformer and developing language-aware neural models. In [20], authors developed a model that divides languages into groups and trains one multilingual model for each group. In [21], authors proposed a training method based on a contrastive learning scheme and data augmentation for a single unified multilingual translation model.

### B. Multilingual NMT with denoising autoencoder

In [22], authors proposed a multilingual unsupervised NMT framework that trains different languages simultaneously with a shared encoder and multiple decoders relying on denoising autoencoding of each language and back-translation between English and numerous non-English languages. In [23], authors presented BART, a denoising autoencoder for pretraining sequence-to-sequence models. BART is trained by corrupting text with random noise and letting the model reconstruct the original text. In [24], authors presented mBART—a sequence-to-sequence denoising auto-encoder pre-trained on large-scale monolingual corpora in many languages using the BART objective. In [25], authors proposed an approach called NMT-Adapt, which combines denoising autoencoding, back-translation and adversarial objectives to utilize monolingual data for low-resource adaptation.

### C. Shortcomings of existing methods

The existing low-resource multilingual translation model (such as [5], [9], [18]) mainly focused on leveraging the high resource language to improve translation quality via a transfer learning approach. There is a need to maintain the coherent latent space between the sentences that helps the model to learn better context. Our approach uses sentence interpolation for generating intermediate latent sentence representation based on adversarial autoencoder and train the multilingual translation model based on Wasserstein-GAN, which optimizes the model performance.

Table II: Symbol description

| Symbol | Description |
|---|---|
| $M$, $Z$ | Sentence space, latent space |
| $m$, $z$ | Sequence, Generated latent variable |
| $B$ | Class labels |
| $C$, $m_{trans}$ | Corpus, Transliterated sequence |
| $E_{RNN}$ | Deterministic encoder in DAAE |
| $D_{RNN}$ | Probabilistic decoder in DAAE |
| $Q_{FFN}$ | Discriminator part of DAAE |
| $\mathcal{L}_{rec}$ | Reconstruction loss in DAAE |
| $\mathcal{L}_{adv}$ | Adversarial loss in DAAE |
| $E(m)$ | Encoded sequence $m$ |
| $m_e$ | Perturbed $m$ |
| $s$, $t$ | Source sentence, Target sentence |
| $s'$, $t'$ | Interpolated source sentence, Interpolated target sentence |
| $z_s$ | Latent variable for source sentence |
| $z_t$ | Latent variable for target sentence |
| $s_{syn}$ | Synthetic source sentence |
| $t_{syn}$ | Synthetic target sentence |
| $t'_{syn}$ | Predicted target sequence |
| $t_{gen}$ | NMT generated sentence |
| $t_{ref}$ | Reference sentence |
| $L_i$ | $i^{th}$ language |
| $\alpha$ | Interpolation factor |
| $\lambda$ | Training hyperparameter in DAAE |
| $h$ | Feature map |
| $G_{NMT}$ | Generator part of translation model |
| $Q_{CNN}$ | Discriminator part of translation model |
| $R_i$ | Reward gained by $i^{th}$ language pair |
| $R_{MMT}$ | Reward for translation model |
| $\theta$ and $\phi$ | Gradient descent for DAAE and DAASI |
| $\beta$ | Interpolation hyperparameter |
| $lr$ | Learning rate |
| $\mathbb{P}_r$ | Probability distribution for human translated sentence |
| $\mathbb{P}_g$ | Probability distribution for machine generated sentence |
| $\mathcal{L}_{gen}$ | Generator loss function in WGAN |
| $\mathcal{L}_{critic}$ | Critic loss function in WGAN |

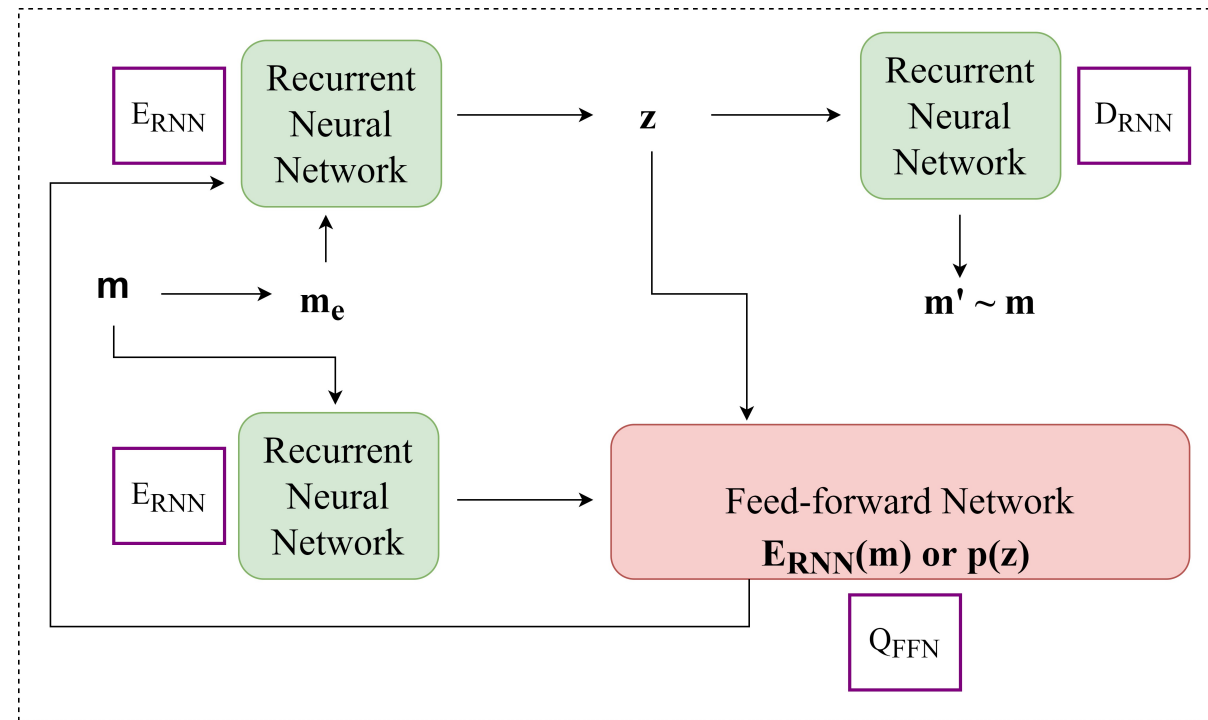

Figure 1: DAAE for multilingual model.

## III. Proposed Approach

This section discusses the proposed DAASI approach, which improves language generalisation in low-resource multilingual machine translation and trains the translation model with modified multilingual-reward via WGAN. DAASI consists of three components: Denoising Adversarial Auto-encoder *(DAAE)* [13], sentence interpolation, and WGAN-based translation model; described in the following:

### A. DAAE for multilingual model

We introduce DAAE (Fig. 1) in the multilingual NMT model for pseudo-corpus generation and encourages the model to implicitly learn a similar latent representation of sentences. DAAE is an encoder-decoder framework consisting of a deterministic encoder $E_{RNN}$, a probabilistic decoder $D_{RNN}$ and a discriminator $Q_{FFN}$. Both $E_{RNN}$ and $D_{RNN}$ are Recurrent Neural Network *(RNN)*. $E_{RNN}$ takes input sequence $m$ and uses the final RNN hidden state as its encoding $z$. $D_{RNN}$ generates $m$ autoregressively. $Q_{FFN}$ is a feed-forward network that calculates the likelihood of $z$ coming from the prior rather than the encoder. Deterministic encoder $E_{RNN}$: $\mathcal{M}\rightarrow\mathcal{Z}$ models data space to latent space. Probabilistic decoder $D_{RNN}$: $\mathcal{Z}\rightarrow\mathcal{M}$ generates sequences from latent representations. Discriminator $Q_{FFN}$: $\mathcal{Z}\rightarrow[0,1]$ attempts to distinguish between encodings of data $E_{RNN}(m)$ and samples from $p(z)$. First, we encode the sequence $m$ from multilingual monolingual corpus into a common roman script via WX-transliteration as follows:

$$m_{trans} = enc_{trans}(m), \tag{1}$$

where $m_{trans}$ and $enc_{trans}$ represent transliterated sequence and WX-encoder, respectively. Then we exploit the transliterated monolingual data by adding noise to each sentence, passing the noised sentence to the encoder, training the model, and recovering the original sentence from the noised sentence using an adversarial autoencoder. We corrupt the data by introducing local m-perturbations in a sequence $m_{trans}$ as follows:

$$m_e = perturb(m_{trans}), \tag{2}$$

where *perturb* and $m_e$ represent perturbation process and the corrupted sequence, respectively.

We employ the two types of losses i.e., reconstruction loss ($\mathcal{L}_{rec}$) and adversarial loss ($\mathcal{L}_{adv}$) to train the DAAE module described as follows [13]:

$$\mathcal{L}_{rec}(\theta_{E_{RNN}}, \theta_{D_{RNN}}) = \mathbb{E}_{p(m,m_e)}\big[-\log p_{D_{RNN}}(m|E_{RNN}(m_e))\big], \tag{3}$$

$$\mathcal{L}_{adv}(\theta_{E_{RNN}}, \theta_{Q_{FFN}}) = \mathbb{E}_{p(z)}\big[-\log Q_{FFN}(z)\big] + \mathbb{E}_{p(m_e)}\big[1-\log Q_{FFN}(E_{RNN}(m_e))\big], \tag{4}$$

where,

$$p(m, m_e) = p_{data}(m)p(m|m_e), \tag{5}$$

$$p(m_e) = \sum_m p(m, m_e). \tag{6}$$

Both reconstruction and adversarial loss are weighted via hyperparameter $\lambda$>0 during training as follows:

$$\min_{E_{RNN},D_{RNN}} \max_{Q_{FFN}} \mathcal{L}_{rec}(\theta_{E_{RNN}}, \theta_{D_{RNN}}) - \lambda\mathcal{L}_{adv}(\theta_{E_{RNN}}, \theta_{Q_{FFN}}). \tag{7}$$

With perturbation process *perturb*, the posterior distributions of the latent representations are of the form:

$$p(z|m_i) = \sum_{m_{e_i}} p_{perturb}(m_{e_i}|m_i)p_{E_{RNN}}(z|m_{e_i}). \tag{8}$$

We describe the DAAE training procedure in the Algorithm 1 for better understanding of the model. First, encode the multilingual monolingual sentence into WX-representation and then corrupt the sentence using the perturbation process (lines 1-2). Then train the $E_{RNN}$ and $D_{RNN}$ by keeping $Q_{FFN}$ fixed (lines 4-6). Sample a batch from corrupted monolingual data and generate $z \sim p(m_{e_i})$ (lines 4-5). Next, we reconstruct $m_i$ from $m_{e_i}$ and compute $\mathcal{L}_{rec}$ via Eq. (3) (line 6). Then train the $Q_{FFN}$ by keeping $E_{RNN}$ and $D_{RNN}$ fixed (lines 7-9). Generate $z \sim p(z|m_i)$ from original data and compute $\mathcal{L}_{adv}$ via Eq. (4) (lines 8-9). Finally, jointly train the $E_{RNN}$, $D_{RNN}$ and $Q_{FFN}$ autoregressively via Eq. (7) until the model gets converged (lines-3-11).

**Algorithm 1:** DAAE training procedure

**Input**: Sentence $(m)$,$\forall m \in$C
**Output**: DAAE-model
1 $m_{trans} \leftarrow enc_{trans}(m)$;
2 $m_e \leftarrow$ perturb$(m_{trans})$;
3 **while** *$\theta$ has not converged* **do**
    `// Update (`$E_{RNN}$`, `$D_{RNN}$`) and keep `$Q_{FFN}$` fixed`
4   sample $\{m_{e_i}\}_{i=1}^{j}$; a batch from corrupted monolingual data
5   Generate $z \sim p(m_{e_i})$;
6   Reconstruct $m_i$ from $m_{e_i}$ and compute $\mathcal{L}_{rec}$ via Eq. (3);
    `// Update (`$E_{RNN}$`, `$Q_{FFN}$`) and keep `$D_{RNN}$` fixed`
7   sample $\{m_i\}_{i=1}^{j}$; a batch from monolingual data
8   Generate $z \sim p(z|m_i)$;
9   Compute $\mathcal{L}_{adv}$ via Eq. (4);
10  Perform min-max using Eq. (7); `// Train the DAAE via min-max algorithm`
11  $\theta \leftarrow \theta$ - lr$\frac{\partial \mathcal{L}_{rec}}{\partial \theta}$; `// Update the parameter`

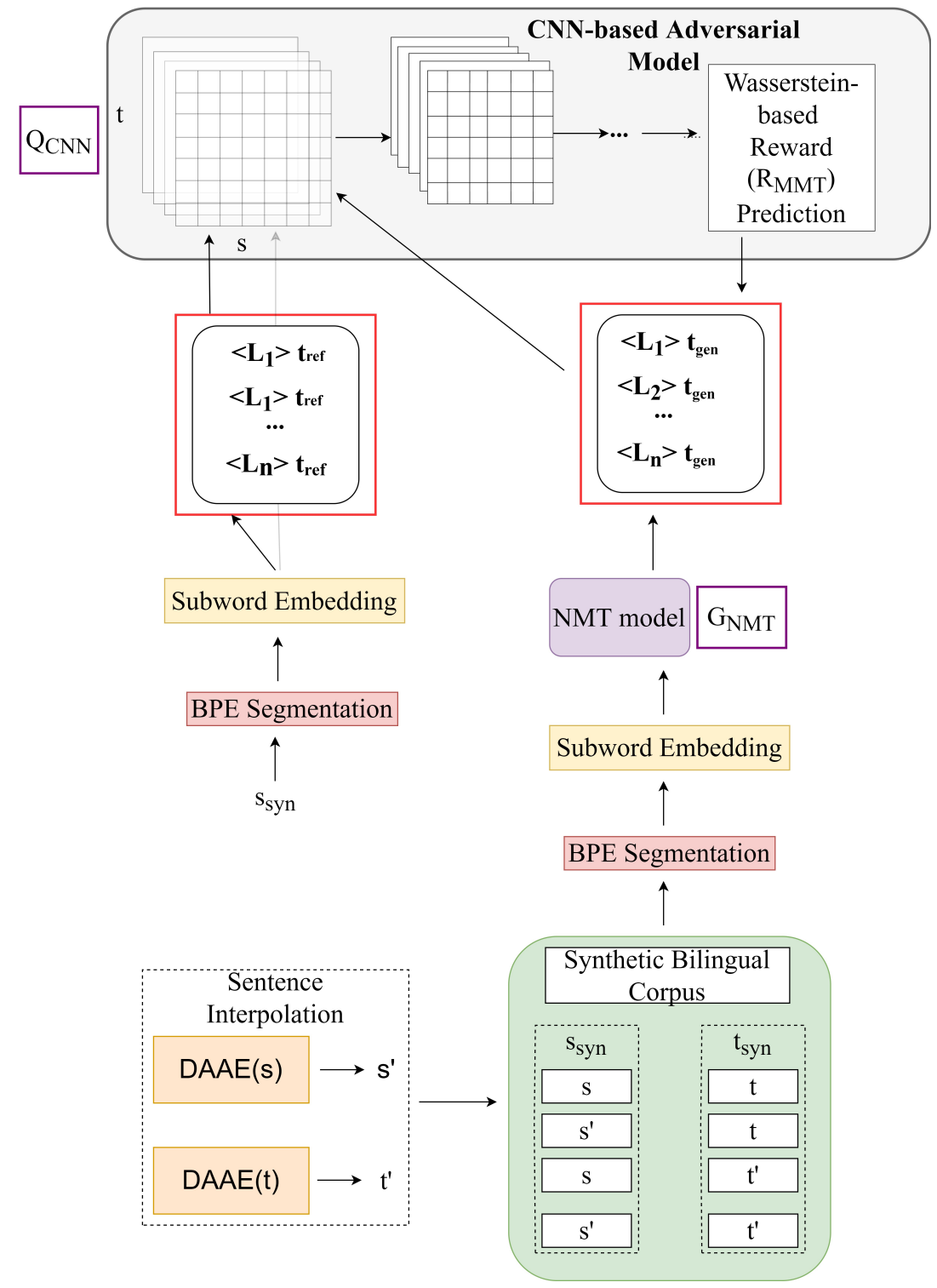

Figure 2: Illustration of proposed architecture.

## B. Sentence interpolation

We perform sentence interpolation between source $s$ and target $t$ language pairs by traversing the latent space of the text auto-encoder. DAASI encodes the source-target language pair to $z_s$ and $z_t$, and decode from $\alpha z_s+(1-\alpha)z_t$ $(0\leq\alpha\leq1)$ into interpolated form $s'$ or $t'$. We perform sentence interpolation on both the source and target monolingual data. To generate $s'$, we train the DAAE on source-side monolingual data and perform the sentence interpolation between source and target parallel data. After performing interpolation, we get many sentences as output. We select the sentence with a high degree of similarity to the source sentence based on the chrF2 [28] measure. Similarly, we train the DAAE on target side monolingual data and perform the sentence interpolation between source and target parallel data to get $t'$. We merge the generated corpus with original training data to create synthetic parallel corpus ($s_{syn}$,$t_{syn}$) for further training of model as shown in Fig. 2. The generated corpus is semantically similar to the original one but contains more information that helps the model learn better context between sentences. We use the sentence interpolation in DAASI to generate the synthetic sentences close to both source and target sentences. These give some intermediate latent representations of sentences that are beneficial for learning the context of relatedness between the languages. Denoising helps produce higher-quality sentence interpolations, suggesting better linguistic continuity in its latent space.

### C. Translation model

We have trained the translation model on the generated semantic parallel corpora ($s_{syn}$,$t_{syn}$) using WGAN, as shown in Fig. 2. It is made up of generator ($G_{NMT}$) and critic ($Q_{CNN}$) parts that use the NMT and Convolutional Neural Network (CNN) architectures, respectively [29]. The generator objective minimises the Wasserstein distance between data distribution of machine-generated translated sentences and human translated references. The Wasserstein distance is a distance metric between two probability distributions on a given metric space. Mathematically, we define the Wasserstein distance $W$ for the translation model between the probability distributions of machine-generated translated sentences $\mathbb{P}_g$ and the human translated references $\mathbb{P}_r$ ($\mathbb{P}_r$ and $\mathbb{P}_g$ belong to embedding space $\mathcal{X}$) as follows [15]:

$$W(\mathbb{P}_r, \mathbb{P}_g) = \inf_{\gamma \in \prod(\mathbb{P}_r, \mathbb{P}_g)} \mathbb{E}_{(t'_{syn}, t_{syn}) \sim \gamma}[||t'_{syn} - t_{syn}||], \quad (9)$$

where $\prod(\mathbb{P}_r, \mathbb{P}_g)$ represents the set of all joint distributions over human translated references $t_{syn}$ and machine translated sentences $t'_{syn}$ such that the marginal distributions are equal to $\mathbb{P}_r$ and $\mathbb{P}_g$, and $\gamma(t'_{syn}, t_{syn})$ is the distance that must be moved from $t'_{syn}$ to $t_{syn}$ to transform $\mathbb{P}_r$ to $\mathbb{P}_g$, respectively. In generator, we use the NMT architecture to train the model but employ the following objective function instead of cross-entropy loss:

$$\begin{aligned} L_{gen}(w) = \min_{\phi} \mathbb{E}_{t_{syn} \sim \mathbb{P}_r}[Q_{CNN}(t_{syn})] \\ - \mathbb{E}_{t'_{syn} \sim p(t'_{syn})}[Q_{CNN}(t'_{syn})]. \end{aligned} \quad (10)$$

In the second part of the model (critic), the objective is to estimate the Wasserstein distance between data distribution of machine-generated translated sentences and human-translated references. For the critic part, we create an image-like representation $h^{(0)}$ by simply appending the embedding vectors of words in $s_{syn}$ and $t_{syn}$ based on [14]. Therefore, for $i^{th}$ word $u_i$ in the source sentence $s_{syn}$ and $j^{th}$ word $v_j$ in the target sentence $t_{syn}$, we have the following feature map [14]:

$$h^{(0)}_{i,j} = [u_i^T, v_j^T]^T. \quad (11)$$

Based on such an image-like representation, we perform convolution on each 3 X 3 window in order to capture the correspondence between segments in $s_{syn}$ and segments in $t_{syn}$ using the following feature map of type $f$:

$$h^{(1,f)}_{i,j} = \sigma(W^{(1,f)} \hat{h}^{(0)}_{i,j} + b^{(1,f)}). \quad (12)$$

where $\hat{h}^{(0)}_{i,j} = [h^{(0)}_{i-1:i+1, j-1:j+1}]$ is the $3 \times 3$ window and $\sigma(x) = \frac{1}{(1+exp(-x))}$ represents sigmoid function. After that we perform a max-pooling in non-overlapping $2 \times 2$ window:

$$h^{(2,f)}_{i,j} = \max(h^{(1,f)}_{2i-1,2j-1}, h^{(1,f)}_{2i-1,2j}, h^{(1,f)}_{2i,2j-1}, h^{(1,f)}_{2i,2j}). \quad (13)$$

This type of 2D architecture helps in modelling the semantic relationship between the two sentences more accurately. Based on this 2D architecture, we train the critic part using the objective function described as follows:

$$\begin{aligned} \mathcal{L}_{critic} = \max_{w \in W} \mathbb{E}_{t_{syn} \sim \mathbb{P}_r}[Q_{CNN}(t_{syn})] \\ - \mathbb{E}_{t'_{syn} \sim p(t'_{syn})}[Q_{CNN}(t'_{syn})]. \end{aligned} \quad (14)$$

$Q_{CNN}$ is updated synchronously with $G_{NMT}$ during model training. We have defined the training for translation model as follows:

$$\begin{aligned} \min_{G_{NMT}} \max_{Q_{CNN}} \mathbb{E}_{t_{syn} \sim \mathbb{P}_r}[Q_{CNN}(t_{syn})] \\ - \mathbb{E}_{t'_{syn} \sim p(t'_{syn})}[Q_{CNN}(t'_{syn})]. \end{aligned} \quad (15)$$

### D. Multilingual reward

In multilingual-NMT, we attach the <*SRC*> tag at the beginning of each source sentence, merge parallel corpora from different languages, and train the translation model. However, in GAN, for the multilingual model, using the merged way of reference translation in different languages for reward computation may negatively affect the model's performance due to the diversity of languages. To handle this problem of language diversity, we have proposed a multilingual reward based on the Wasserstein distance between the sentences, as shown in Fig. 3. Given a synthetic source sentence $s_i$, a critic based on CNN is proposed to distinguish between the model translation results $t_{gen}$ and the reference translation $t_{ref}$. The translation matching of a (source, target) sentence pair must be measured to accomplish this task. We use the Wasserstein distance between real and machine-generated sentences as the objective of reward measurement in one language. The objective $R_i$ calculates the reward for $i^{th}$ language pair, which measures the Wasserstein distance between real $t_{ref}$ and generated sentence $t_{gen}$ as follows:

$$R_i = \inf_{\gamma \in \prod(P_r, P_g)} \mathbb{E}_{(t_{gen}, t_{ref}) \sim \gamma}[||{t^i}_{gen} - {t^i}_{ref}||]. \quad (16)$$

The presence of multiple morphologically rich languages in multilingual NMT impedes the generator's ability to learn effective parameters from WGAN model training. To effectively train the WGAN-based multilingual model, we extend the critic of WGAN for MMT by incorporating multiple references of different language pairs for reward calculation during model training. For reward computation in the multilingual-NMT model, we divide the test set language pairs into $K$ sets such that each set has pairs of different source languages. For a better-optimized model, we compute the reward on the same number of different language pairs. We compute the reward $R_{MMT}$ for multilingual-GAN as follows:

$$R_{MMT} = \frac{\sum_{i=L_1}^{L_n} \inf_{\gamma \in \prod(P_r, P_g)} \mathbb{E}_{(t_{gen}, t_{ref}) \sim \gamma}[||{t^i}_{gen} - {t^i}_{ref}||]}{n}, \quad (17)$$

where, $n$ is the total number of language pairs in one set of $K$.

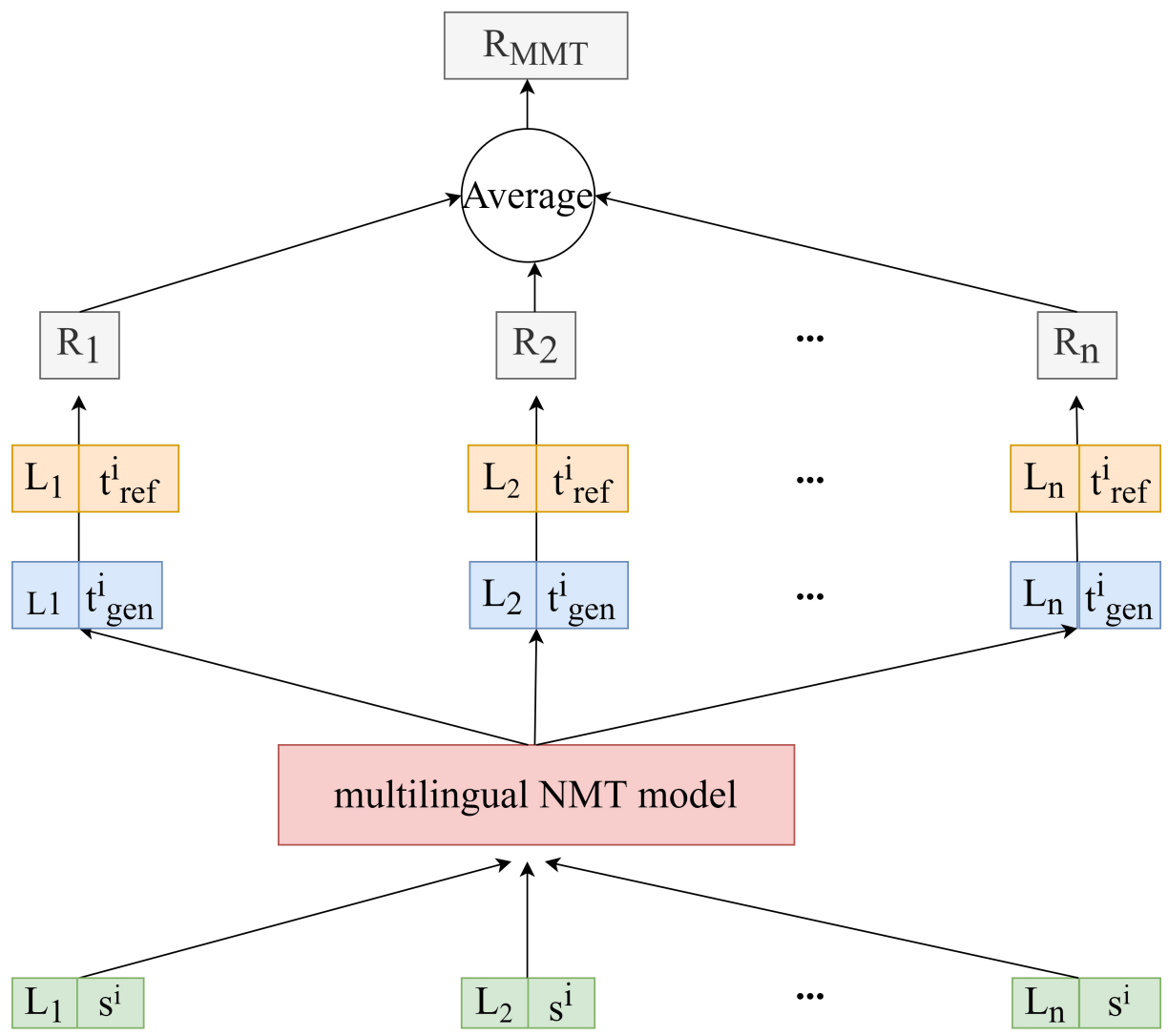

Figure 3: Multilingual reward generation.

**Algorithm 2:** DAASI training procedure

**Input**: Parallel($s$,$t$) $\forall$ $s$ $\in$Source language, $\forall$ $t$ $\in$Target language, $\rho$ = 0.00005, $c$ = 0.01, $n_{critic}$ = 5, $G_{NMT}$ and $Q_{CNN}$ with parametric function denoted as $G_\phi$ and $Q_w$, respectively.

**Output**: DAASI-based multilingual NMT model ($G_\phi$)

1 DAAE($s$) and DAAE($t$);
2 s' $\leftarrow$ Interpolate($s$,$t$) via DAAE($s$); `// Perform interpolation on source.`
3 t' $\leftarrow$ Interpolate($s$,$t$) via DAAE($t$); `// Perform interpolation on target.`
4 $\{s_{syn}, t_{syn}\}$ $\leftarrow$ merge($\{s', t\}$, $\{s, t\}$, $\{s, t'\}$, $\{s', t'\}$);
5 Pretrain $G_\phi$ and $Q_w$ on real data;
6 **while** *$\phi$ has not converged* **do**
7   **for** $b$ = 0, ..., $n_{critic}$ **do**
8     sample $\{{t_{syn}}^i\}_{i=1}^n \sim \mathbb{P}_r$; `// a batch from synthetic data`
9     sample $\{{t'_{syn}}^i\}_{i=1}^n \sim p(t_{syn}|s_{syn})$; `// a batch of machine generated samples`
10     $G_w \leftarrow \nabla_w$ [$\frac{1}{n}\sum_{i=1}^n Q_w({t_{syn}}^i)$ - $\frac{1}{n}\sum_{i=1}^n Q_w(G_\phi({t'_{syn}}^i))$];
11     Compute $R_{MMT}$;
12     **if** $R_{MMT_b} \leq R_{MMT_{b-1}}$ **then**
13       return $G_\phi$;
14     $w \leftarrow w + \rho$ . RMSProp($w$,$G_w$); `// Update the critic parameters`
15     $w \leftarrow$ clip(w,-c,c);
16   sample $\{{t'_{syn}}^i\}_{i=1}^n \sim p(t_{syn}|s_{syn})$; `// a batch of machine generated samples`
17   $G_\phi \leftarrow - \nabla_\phi \frac{1}{n}\sum_{i=1}^n Q_w(G_\phi({t'_{syn}}^i))$;
18   $\phi \leftarrow \phi + \rho$ . RMSProp($\phi$,$G_\phi$); `// Update the generator parameters`

### E. Model training

In order to make our work reproducible, we describe the training details of our proposed approach in Algorithm 2. First, we pretrain the DAAE on the monolingual source $s$ and target $t$ languages using Algorithm 1 (line 1). Then, generate a semantic corpus for parallel source and target sentences via the sentence interpolation approach and merge the generated semantic corpus with the original multilingual training corpus(lines 2-4). Pretrain the $G_\theta$ and $Q_w$ on real training data (line 5). We Update the critic model five times more than the generator for each iteration (lines 7-15). We sample the batch from real data and the machine-generated data for each critic and compute the Wasserstein loss (lines 8-10). Then, compute the reward and if the reward is less than or equal to the reward in the previous iteration, return the generator model; otherwise, update the parameters and perform clipping (lines 11-15). Sample the batch of machine-generated data for the generator and compute the generator's Wasserstein loss (lines 16-17). Finally, update the generator parameters and Train the generator and critics autoregressively until the model gets converged (lines 6-18).

## IV. Performance Study

In this section, we discuss the corpus statistics, experimental setup and result analysis on various parameters to execute the experiments.

### A. Datasets

We evaluate our proposed method on five language pairs in totally 10 directions: Gujarati↔Hindi, Nepali↔Hindi, Urdu↔Hindi, Maithili↔Hindi and Punjabi↔Hindi. We also evaluate our method on zero-shot language pairs (Bhojpuri↔Hindi and Magahi↔Hindi) under unsupervised scenario. For *Gujarati↔Hindi* translation task, we extract the bilingual datasets from CVIT-PIB which consists of 15000 training, 1973 test and 1973 validation sentence pairs [30]. For *Nepali↔Hindi* translation task, training, test, and validation data of about 133000, 3000, 3000 respectively collected from WMT19 task [31], Opus [32], and TDIL repositories [33]. For *Punjabi↔Hindi*, *Maithili↔Hindi*, and *Urdu↔Hindi*, we gather the bilingual data from Opus only [32]. For unsupervised experiments, test set of Bhojpuri↔ Hindi and Magahi ↔ Hindi are collected from zero-shot translation task at LoResMT2020 [34]. In Punjabi↔Hindi and Maithili↔Hindi translation task, we use 200000 and 93000 training sentences for Punjabi↔Hindi and Maithili↔Hindi respectively. Testing and validation are done on 7000 sentences for Punjabi↔Hindi and 3000 sentences for Maithili↔Hindi translation tasks. For Urdu↔Hindi, we train the model on 100000 sentences. Testing and validation are done on 3000 sentences.

### B. Settings

In this section, we discuss the different framework and hyper-parameter used to train the models. Our proposed DAASI approach consists of two components: DAAE and a Multilingual-based reward for the NMT model. We have trained DAAE using the framework and settings discussed in [13]. For multilingual NMT, we have built a critic model based on CNN as described in [14], and for generator, we use a transformer based on Fairseq framework [35], by changing the loss function with Wasserstein loss and executed the experiments on PARAM Shivay supercomputer with an Nvidia V100 GPU. We pre-process the data with *SentencePiece* library [36]. Our model trained on the 5 number of the decoder and encoder layers with a 512 embedding dimension for each and learns the joint vocabulary by sharing dictionary and embedding space. The feed-forward network has encoder and decoder embedding dimensions equal to 2048. The number of attention heads used for the decoder and encoder is 2. We use weight decay, label smoothing and dropout for regularisation, with the corresponding hyper-parameters to 0.0001, 0.2 and 0.4, respectively. We use Adam optimizer for the model having $\beta 1 = 0.9$ and $\beta 2 = 0.98$, and keep the *patience* value equal to 10.

Table III: Results on X → Hindi

| Model | GU → HI | NE → HI | PA → HI | MAI → HI | UR → HI |
|---|---|---|---|---|---|
| **LSTM** | 21.1 | 26.9 | 57.6 | 62.3 | 13.1 |
| **Transformer-NMT** | 20.8 | 28.5 | 57.1 | 64.3 | 13.4 |
| **Adversarial-NMT** | 21.3 | 28.9 | 58.5 | 65.1 | 13.1 |
| **DAAE+NMT** | 21.7 | 29.6 | 58.8 | 65.1 | 14.5 |
| **DAASI** | 24.2 | 32.1 | 59.2 | 65.9 | 17.4 |

Table IV: Results on Hindi → X

| Model | HI → GU | HI → NE | HI → PA | HI → MAI | HI → UR |
|---|---|---|---|---|---|
| **LSTM** | 20.6 | 31.0 | 58.7 | 64.1 | 11.6 |
| **Transformer-NMT** | 21.8 | 30.1 | 59.3 | 66.2 | 12.5 |
| **Adversarial-NMT** | 22.2 | 30.8 | 60.6 | 66.9 | 12.7 |
| **DAAE+NMT** | 23.1 | 31.2 | 60.9 | 66.8 | 13.2 |
| **DAASI** | 26.4 | 34.5 | 62.1 | 68.1 | 14.8 |

To demonstrate the effectiveness of our approach, we compare with the following baseline models:

*a) LSTM and Transformer:* We have trained the baseline NMT model with the LSTM and Transformer architecture using Fairseq, a sequence modelling toolkit and executed the experiments on an Nvidia V100 GPU [35].

*b) Adversarial-NMT:* We have trained the Adversarial-based NMT model using the architecture and settings described in [14].

### *C. Results and analysis*

To measure the performance of our proposed model, we use BLEU [37] as evaluation metric. This evaluation metric judges the model's performance by focusing on semantic, syntactic, morphological and fluency factors of generated hypothesis. In the following section, we describe the results and analyze the proposed approach under multilingual and unsupervised conditions.

*1) Multilingual effect of different methods:* Tables III and IV contain the results of different methods performed under multilingual settings. Compared with the traditional multilingual NMT model (LSTM and Transformer), the Adversarial-based multilingual NMT approach achieved better performance due to its advantage on optimisation of training objective of NMT- to force the translation results to be as similar as ground-truth translation generated by a human. However, due to its optimisation limit, the Adversarial-NMT method failed to outperform the DAAE-based NMT. DAAE-based NMT, in addition to adversarial, works better due to its de-corrupting text features. Therefore, it outperforms the Adversarial-based multilingual NMT method. Existing methods only take de-corrupting text as input features in model training, which could not better model contexts for rich morphology languages. In addition, existing models only apply a single reference when distinguishing between model-generated translation and human translation. Our proposed model incorporating sentence interpolation with DAAE features for generating sentences with better context for morphological languages and multiple reference translation in the existing best multilingual-NMT model achieved the best performance of upto 4 BLEU points on all translation language tasks.

*2) Effect of different components:* The proposed DAASI model consists of DAAE-based sentence interpolation and GAN-based multilingual training. From Tables III and IV, we have observed that removing any components hurt the model performance with a sufficient gap in evaluation scores. Using only denoising auto-encoder gives good results compared to vanilla NMT, but adding sentence interpolation improves the model performance. The reason behind such performance gain is the better context learning of the model by sentence interpolation approach. We have also observed that using a multilingual reward for training the model improves the model performance up to 3 BLUE points. The proposed DAASI approach achieved the best performance, incorporating both features from two components.

*3) Effect of language similarity:* We compute the perplexity score of languages to assess their similarity to one another. Each language has been trained and tested for perplexity. The perplexity of each language corpus is computed as follows:

$$PP(C) = \sqrt[n]{\frac{1}{P(x_1, x_2, ..., x_n)}}, \tag{18}$$

where $P(x_1, x_2, ..., x_n)$ is probability of a sequence of sentences$\{x_1,x_2,...,x_n\}$ in a corpus $C$, computed as follows:

$$P(x_1, x_2, ..., x_n) = \prod_{i=1}^{m} p(x_i). \tag{19}$$

Table V lists the perplexity-based scores of demonstrated languages with each other. The values in Table V indicate how closely languages are related to one another. Languages with lower perplexity scores between them are more similar to each other. The degree of similarity between languages decreases as perplexity increases. This similarity between languages empirically justifies the relatedness between languages that show highly co-relation with the results obtained in Tables III and IV. We have observed that languages having better similarity score between them perform better. For example, NE↔HI shows wide range of improvement compared to other language pairs.

Table V: Perplexity

| **Score** | **GU** | **NE** | **HI** | **PA** | **MAI** | **UR** |
|---|---|---|---|---|---|---|
| **GU** | 3.144 | 229.256 | 39.369 | 80.830 | 91.802 | 332.139 |
| **NE** | 191.945 | 3.528 | 170.825 | 263.506 | 121.894 | 439.234 |
| **HI** | 38.027 | 222.341 | 3.915 | 55.358 | 94.243 | 367.443 |
| **PA** | 113.167 | 354.254 | 75.479 | 3.526 | 113.607 | 406.797 |
| **MAI** | 121.656 | 226.242 | 112.471 | 115.610 | 3.225 | 498.948 |
| **UR** | 380.829 | 777.826 | 649.583 | 409.671 | 242.704 | 3.317 |

*4) Effect of morphological complexity between languages:* Our study primarily includes morphologically diverse languages. To correlate our findings with the morphological richness of languages, we have used corpus-based complexity scales as discussed below.

*a) Word-entropy of languages:* The average information content of words is represented by Entropy. This metric would be higher in languages with a greater variety of word forms, i.e. languages that acquire more information into word structure rather than phrase or sentence structure.

Let $C$ be a text drawn from a vocabulary $V = \{v_1, v_2, \ldots, v_k\}$ of size $k$. Furthermore, let word type probabilities are distributed according to $p(v) = P_r(v \in C)$ for $v \in Z$. The average information content of the word types is calculated by Shannon [38] method as follows:

$$H(C) = -\sum_{j=1}^{k} p(v_j) \log_2(p(v_j)). \tag{20}$$

*b) Type-to-Token Ratio (TTR) of languages:* To calculate morphological complexity, we consider the ratio of word types over word tokens [39]. The spectrum of word forms is expanded by using productive morphological markers. As a result, higher TTR value implies higher morphological complexity. Given a text $C$ drawn from a vocabulary of word types $V = \{v_1, v_2, \ldots, v_k\}$, the measure is written as follows:

$$TTR(C) = \frac{k}{\sum_{j=1}^{k} f(q_j)}, \tag{21}$$

where, $f(q_j)$ is the token frequency of the $j^{th}$ type.

Entropy and TTR with higher values indicate language having high lexical richness as shown in Table VI. We have observed that translation of language pairs from low to high morphological complexity gives better score than high to low. For example, GU→HI gives better BLEU scores than HI→GU language pairs.

Table VI: Entropy and TTR

| Languages | Entropy | TTR |
|---|---|---|
| HI | 3.797 | 3.936e-05 |
| GU | 3.745 | 3.141e-05 |
| NE | 3.577 | 5.260e-05 |
| PA | 3.793 | 1.773e-05 |
| MAI | 3.916 | 1.078e-4 |
| UR | 4.119 | 9.883e-06 |

*5) Using multilingual model for zero-shot language pairs:* Tables VII and VIII list the result of experiments performed under unsupervised settings. For evaluating the model on unsupervised conditions, we demonstrate the experiments on BHO↔HI and MAG↔HI zero-shot language pairs. We use the multilingual pretrained model to evaluate the results on zero-shot language pairs. The reason for opting these language pairs is closely relatedness between all the multilingual language pairs and the zero-shot language pairs. Our models under unsupervised conditions also succeed in improving upto 5 BLEU points. The reason behind better performance is that the speakers of these closely related languages are crossing the borders for a long period of time, leading to the sharing of linguistic and phonetic features between the languages.

Table VII: Unsupervised results on X → Hindi

| **Model** | **BHO → HI** | **MAG → HI** |
|---|---|---|
| **LSTM** | 16.7 | 11.0 |
| **Transformer-NMT** | 19.5 | 13.7 |
| **Adversarial-NMT** | 20.1 | 15.1 |
| **DAAE+NMT** | 20.5 | 16.3 |
| **DAASI** | 22.1 | 17.9 |

Table VIII: Unsupervised results on Hindi → X

| **Model** | **HI → BHO** | **HI → MAG** |
|---|---|---|
| **LSTM** | 2.5 | 2.9 |
| **Transformer-NMT** | 2.6 | 3.1 |
| **Adversarial-NMT** | 2.8 | 3.7 |
| **DAAE+NMT** | 3.4 | 3.9 |
| **DAASI** | 4.2 | 4.8 |

*6) Translated examples:* We have translated some sentences from different language pairs using a different method and presented them in Fig. 4. Here, we include the examples of Gujarati→Hindi and Nepali→Hindi translation pairs. Ongoing through examples, we see that predicted sentence pairs in our model gave a more semantic representation of sentences than the existing baseline method. LSTM and Transformer versions of translations were not good enough to represent the meaning of the source sentences. Whereas, Adversarial-NMT and DAAE-based NMT shown improvement in translated sentences compared to reference sentences. However, our proposed approach gave the meaning more closer to the references and shown its effectiveness.

| **Language Pair-1** | **Gujarati→Hindi** |
|---|---|
| Source | સંયુક્તપણે વિજ્ઞાન પરિષદો અને કાર્યશાળાઓનું આયોજન |
| Reference | संयुक्त वैज्ञानिक सम्मेलन और कार्यशालाएं आयोजित की जाएं। |
| LSTM | संयुक्त रूप से सम्मेलनों का आयोजन |
| Transformer | संयुक्त रूप से विज्ञान सम्मेलनों और कार्यशालाओं का आयोजन |
| Adversarial-NMT | संयुक्त रूप से विज्ञान सम्मेलनों और कार्यशालाओं प्रस्तुत की जाए |
| DAAE+NMT | संयुक्त रूप से विज्ञान सम्मेलनों और कार्यशालाओं का आयोजन हो |
| DAASI | संयुक्त रूप से विज्ञान सम्मेलनों और कार्यशालाओं का आयोजन किया जाना चाइये |
| **Language Pair-2** | **Nepali→Hindi** |
| Source | तपाईंले जाँच गर्न चाहनुभएको भाषाहरू विरूद्ध टाइप गरिएको शब्द जाँच गर्ने या नगर्ने। |
| Reference | क्या उन भाषाओं के लिए टाइप किए शब्दों को जाँचना चाहिए जिसके लिए आप जाँचना चाहते हैं |
| LSTM | क्या आप जिन  भाषाओं को जांचना चाहते हैं, उनके शब्द को जांचना है या नहीं |
| Transformer | क्या आप जिन  भाषाओं को जांचना चाहते हैं, उनके शब्द को जांचना है या नहीं |
| Adversarial-NMT | क्या जिन  भाषाओं की जांच करना चाहते हैं, उनके शब्द को जांचना है या नहीं |
| DAAE+NMT | आप जिन भाषाओं की जांच करना चाहते हैं, उनके शब्द को जांचना है या नहीं |
| DAASI | आप जिन भाषाओं की जांच करना चाहते हैं, उनके सामने टाइप किए गए शब्द को जांचना है या नहीं |

Figure 4: Translated examples of using different methods.

## V. Conclusion

In this paper, we proposed the *DAASI* approach based on denoising adversarial auto-encoder that performed sentence interpolation by learning the intermediate latent representation of the source and target sentence of multilingual language pairs. Apart from denoising the adversarial autoencoder, we also modified the reward for multilingual NMT with WGAN. The experiments performed on Gujarati↔Hindi, Nepali↔Hindi, Punjabi↔Hindi, Maithili↔Hindi and Urdu↔Hindi translation tasks demonstrated the effectiveness of our method. In future, we will work on achieving new state-of-the-art performance for the NMT system by fully exploiting the knowledge representation of languages at different granularity levels.

## Acknowledgment

The support and the resources provided by PARAM Shivay Facility under the National Supercomputing Mission, Government of India at the Indian Institute of Technology, Varanasi are gratefully acknowledged.

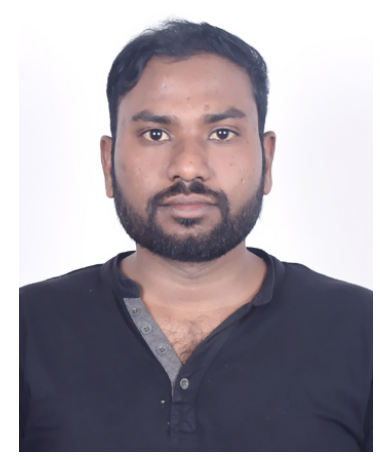

**Amit Kumar** received the M.Tech. degree in computer science and engineering from the National Institute of Technology, Patna, India in 2017. At present, he is a Ph.D. student in the Department of Computer Science and Engineering, Indian Institute of Technology (BHU) Varanasi, India. His research interests include Natural Language Processing, Deep learning, and Artificial Intelligence.

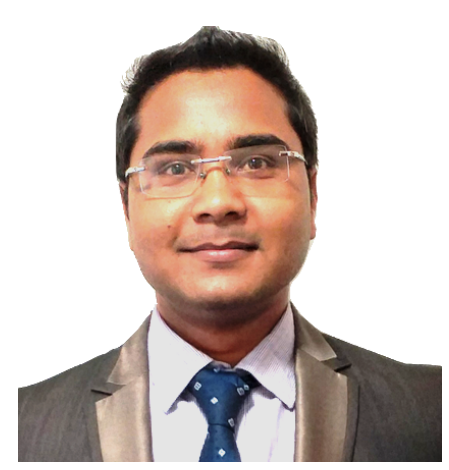

**Ajay Pratap** is an Assistant Professor with the Department of Computer Science and Engineering, Indian Institute of Technology (BHU) Varanasi, India. He worked as a Postdoctoral Researcher in the Department of Computer Science at Missouri University of Science and Technology, USA, for almost 1.5 years.

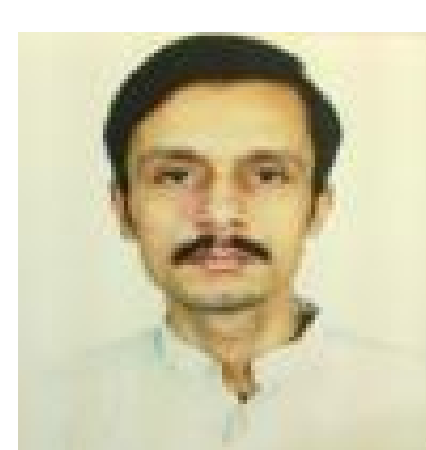

**Anil Kumar Singh** is an Associate professor with the Department of Computer Science and Engineering, Indian Institute of Technology (BHU) Varanasi, India. He worked as a Postdoctoral Researcher in TLP at LIMSI-CNRS, Paris, France, from May, 2012 to April, 2013.